\pdfoutput=1
\documentclass[11pt]{article}

\usepackage[final]{acl}

\usepackage{times}
\usepackage{latexsym}
\usepackage{booktabs}
\usepackage{colortbl} 
\usepackage[normalem]{ulem}
\useunder{\uline}{\ul}{}
\usepackage[T1]{fontenc}
\usepackage[utf8]{inputenc}

\usepackage{microtype}
\usepackage{amsmath}
\usepackage{amsfonts}
\usepackage{url}

\usepackage{inconsolata}

\usepackage{graphicx}

\title{Seeing Through VisualBERT: A Causal Adventure on Memetic Landscapes}

\author{Dibyanayan Bandyopadhyay \\
  Indian Institute of Technology Patna \\
  Patna, Bihar, India \\
  \texttt{dibyanayan\_2321cs14@iitp.ac.in} \\\And
  Mohammed Hasanuzzaman \\
  EEECS, Queen's University Belfast \\
  Belfast, BT7 1NN, UK \\
  \texttt{m.hasanuzzaman@qub.ac.uk} \\\AND
  Asif Ekbal \\
  Indian Institute of Technology Jodhpur \\
  Jodhpur, Rajasthan, India \\
  \texttt{asif@iitp.ac.in} \\
  }

\begin{document}
\maketitle

\begin{abstract}
   Detecting offensive memes is crucial, yet standard deep neural network systems often remain opaque. Various input attribution-based methods attempt to interpret their behavior, but they face challenges with implicitly offensive memes and non-causal attributions. To address these issues, we propose a framework based on a Structural Causal Model (SCM). In this framework, VisualBERT is trained to predict the class of an input meme based on both meme input and causal concepts, allowing for transparent interpretation. Our qualitative evaluation demonstrates the framework's effectiveness in understanding model behavior, particularly in determining whether the model was right due to the right reason, and in identifying reasons behind misclassification. Additionally, quantitative analysis assesses the significance of proposed modelling choices, such as de-confounding, adversarial learning, and dynamic routing, and compares them with input attribution methods. Surprisingly, we find that input attribution methods do not guarantee causality within our framework, raising questions about their reliability in safety-critical applications\footnote{\textcolor{red}{This paper contains various racist and offensive memes and keywords which do not reflect authors' beliefs.}}. The project page is at: \url{https://newcodevelop.github.io/causality_adventure/}
\end{abstract}

\section{Introduction}
Memes have evolved from spreading humor to being used for disseminating offensive content, necessitating the development of neural multimodal systems to detect such content~\cite{kiela2021hateful}. However, these systems often lack transparency, undermining public trust in real-world applications and highlighting the need for interpretability and trustworthiness.

While Large Language Models (LLMs) and Vision Language Models (VLMs) could predict offensive memes and provide self-explanations, these explanations are not always faithful to model behavior~\cite{madsen2024selfexplanations, agarwal2024faithfulness}. Our self-consistency checks (\S Appendix \ref{scc}) confirm this issue for offensive meme detection, prompting us to focus on enhancing the reliability of existing multimodal classifiers.
\begin{figure}
    \centering
    \includegraphics[width=\columnwidth]{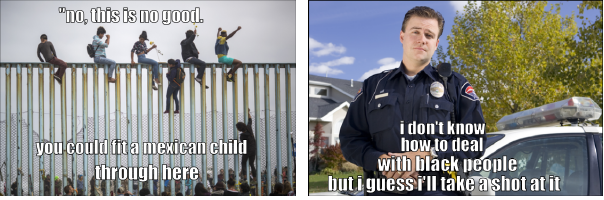}
    \caption{The Underlying notion of these memes is \textit{racism}, which is implicit. None of the input attribution methods could decipher the notion of \textit{racism} solely through input attribution.}
    \label{fig:ip-fail}
\end{figure}
Current interpretability techniques, such as input attributions (e.g., Integrated Gradient \cite{sundararajan2017axiomatic}), struggle with implicit content and causality. For implicitly offensive memes, attribution methods fail to capture underlying concepts like `racism' (refer to Figure \ref{fig:ip-fail}). Also, they only detect influential features without considering their causal impact \cite{chattopadhyay2019neural}.

Causality-based techniques like CausaLM~\cite{feder-etal-2021-causalm} and Amnesic Probing~\cite{elazar2021amnesic} offer solutions but face scalability issues and focus on global rather than local explanations like: `Is adjective important for sentiment analysis' or `Is part-of-speech information crucial for word prediction?' We address these limitations with a novel causal framework integrating VisualBERT~\cite{li2019visualbert} with causal concepts. Our approach extracts implicit context from the meme as a set of causal concepts and uses dynamic routing and adversarial learning to predict meme offensiveness based on both meme content and contribution from causal concepts.

Quantitative analysis (Section \ref{quant}) shows that traditional interpretability techniques, which rely on correlation, do not always align with causality, emphasizing that \textit{correlation does not imply causation}. Through ablation studies, we demonstrate that our framework bases its predictions on relevant causal concepts, enhancing trustworthiness. Qualitative analysis (Section \ref{quals}) indicates whether the model's decisions are justifiable and explains its error cases.

Our proposed framework is novel, model-agnostic, and acts as a proof-of-concept which demonstrates the potential of using causal analysis to elucidate the decision-making process of multimodal classifiers.

\section{Related Work}

\textbf{Causal Interpretability:} Causal interpretability aims to understand how counterfactuals cause model outputs to change, thus estimating the causal effect of inputs~\cite{feder-etal-2022-causal}. A subfield, causal mediation analysis, explores the mechanisms behind these effects~\cite{geiger2021causal,vig2020causal,meng2023locating}. Generating exact counterfactuals is challenging~\cite{abraham2022cebab,calderon-etal-2022-docogen}, so recent work focuses on approximations~\cite{geiger2021causal} or counterfactual representation~\cite{feder-etal-2021-causalm,elazar2021amnesic,ravfogel-etal-2021-counterfactual}. Our current research concentrates on counterfactual representation. Most of the existing works target single modality (e.g. text or vision)~\cite{feder-etal-2021-causalm,goyal2020explaining} and answer global questions about feature importance~\cite{elazar2021amnesic}. We propose a method for answering local questions about specific concepts (e.g., `Is the meme offensive due to the presence of \textit{Holocaust} as a concept?') while addressing scalability issues of prior methods~\cite{feder-etal-2022-causal}. Our framework incorporates concept annotations and integrates with VisualBERT for trustworthy local causal interpretability.

\textbf{Multimodal Interpretability.} Recently, there has been a surge in multimodal models for various tasks~\cite{ding2021cogview,du2022glm,li2023blip2,liu2023visual,zhu2023minigpt}, yet research on generating explanations for their predictions remains limited. Researchers primarily rely on interpretability techniques like LIME~\citep{ribeiro2016why} and SHAP ~\citep{lundberg2017unified} and various input attribution methods~\cite{sundararajan2017axiomatic, 10.5555/3295222.3295230, shrikumar2017just}. However, recently, there has been a shift towards generating natural language explanations, bridging the gap between text-only and multimodal systems. Methods like NLX-GPT~\citep{sammani2022nlxgpt} and Semantify~\citep{ijcai2024p684} offer solutions but fail to fully capture implicit causal meanings or the causal impact of input features~\cite{chattopadhyay2019neural}. This gap motivated us to develop a framework that enables causal interpretations of implicit inputs. 

% Causal interpretability constitutes a multifaceted domain focused on understanding the causal relationships between input features and model outputs. Central to this field is the exploration of counterfactuals, wherein the effect of altering input values on model predictions is investigated. Additionally, causal mediation analysis, situated within mechanistic interpretability, further enriches this domain. The explicit generation of counterfactuals poses a significant challenge, particularly within the domains of natural language processing (NLP) and multimodal AI. Consequently, efforts have shifted towards approximating counterfactuals or generating counterfactual representations. This paper adopts the latter approach, emphasizing counterfactual generation. Most of the works in this realm are in unimodal domains (either NLP\cite{causalm} or in vision\cite{icace}). Also, most causal interpretability ideas answer question like ``Is adjective important for sentiment analysis"\cite{causalm} or ``Is part-of-speech information crucial for word prediction?"\cite{amnesic probing}, while we want to answer questions like ``Is the input meme offensive due to the presence of the concept Holocaust"? Although local causal explainability methods exist\cite{cxplain}, they suffer from extreme scalability issues as the number of concepts increases. Our proposed framework is imbibed with both concept annotations and its seamless integration with VisualBERT, thus making local causal interpretability easy and making the model more trustworthy.

\section{Causal Process}\label{dgp}
\subsection{Causal Diagram}\label{caudiag}

\begin{figure}[ht]
    
    \includegraphics[width=\columnwidth]{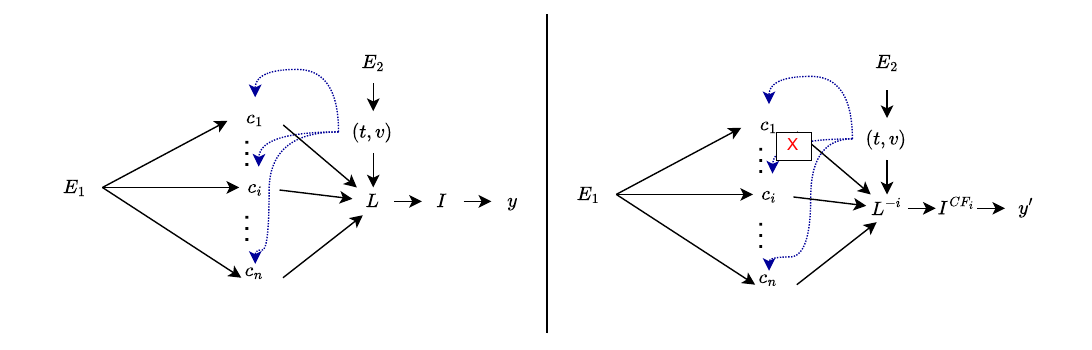}
    \caption{\textbf{Left:} The causal process is illustrated by a SCM. \textbf{Right:} Causal intervention selectively intervenes on a concept $c_i$ to nullify its effect on the model. This generates the intermediate counterfactual representation $I^{CF_i}$. To measure the causal effect of concept $c_i$, we take the Individual treatment effect (ITE) as $|y'-y|$. Dotted blue lines denote that meme content representations ($t$,$v$) (along with $E_1$) generate causal concepts $c_i$. \textcolor{red}{X} demonstrates causal intervention, i.e. breaking the link between ($t$,$v$) and $c_i$, realized by setting $w_i=0$.}
    \label{fig:el}
\end{figure}

Our framework is based on a Structural Causal Model (SCM) that integrates both the causal explanation process and multimodal classification objectives, drawing inspiration from \citet{geiger2021causal}. We assume an exogenous variable $E_1$ that generates causal concepts $c_1$, ... to $c_n$. Another exogenous variable $E_2$ controls meme text $t$ and image $v$ representation. The collection of the concepts $\{c_i\}_{i=1}^n$ controls the latent representation $L$. \textit{$(t,v)$ along with $L$ controls the intermediate representation $I$, which further controls the output $y$ of the model.} Figure \ref{fig:el} represents this in details.

% Following CausaLM (\textit{page 345 defn 3, last 4-5 lines, page 344 last 4 lines, page 336 last para, page 342 paragraph Counterfactual Text Representation}), we want to intervene $I$ such that the generated counterfactual representation $I^{CF_i}$ is insensitive to concept $c_i$ and similar to $I$ in all other concepts $\{c_1,....c_n\} - \{c_i\}$.

\textbf{Counterfactual Representation.} Inspired by CausaLM~\cite{feder-etal-2021-causalm}, we want to intervene $I$ such that the generated counterfactual representation $I^{CF_i}$ is insensitive to concept $c_i$ and similar to $I$ for all the other concepts except $c_i$\footnote{Note a slight abuse of notation here, $c_i \in \mathbb{R}^{1 \times 768}$ refers to concept representation instead of `textual' concepts.}.

% To achieve this, we observe the SCM in Figure \ref{fig:el} where $I$ is a function of $L$ and $(t,v)$, $I = f(L, (t,v))$ where $L$ is known as concept latent which is a weighted sum of every concept representation $c_i$, such that $L = \sum_{i=1}^{n} w_i \cdot c_i$. Now creating the counterfactual representation that is not affected by concept $c_i$ is very easy, just setting $w_i=0$, would suffice. This explicit modelling possesses the benefits of i) simple implementation, ii) $I^{CF}_i$ represents a counterfactual as it is unaffected by $c_i$, whereas affected by every other concept as $I$. 

To achieve this, we observe the SCM, depicted in Figure \ref{fig:el}, where  $I = f(L, (t,v))$ and $L$ represents the latent, a weighted sum of concept representations $c_i$, given by $L = \sum_{i=1}^{n} w_i \cdot c_i$. Creating a counterfactual latent ($L^{-i}$) unaffected by a specific concept $c_i$ is straightforward to achieve by setting $w_i=0$. This explicit modelling offers simple implementation and ensures $I^{CF_i}$ represents a counterfactual unaffected by $c_i$, while still influenced by other concepts.

\textbf{Need for De-confounding.} In this formulation of obtaining the counterfactual representation, a challenge arises because concept $c_i$ might be correlated with other concepts $\{c_j\}$ (e.g. `holocaust' is correlated to `nazism'), such that $L^{-i}$ can be retrieved from $\{c_j\}$. We term these influencing concepts like $\{c_j\}$ as `control concepts' and $c_i$ as the `treated concept', following the terminology in \citet{feder-etal-2021-causalm}. To address this challenge, we propose a novel step called de-confounding (refer to Section \ref{dec}). Here, we constrain our framework to prevent the recovery of a treated concept $c_i$ from the counterfactual latent representation $L^{-i}$, which is essentially a linear combination of potential control concepts. The intuition is that once a concept is removed from the representation containing it, it cannot be recovered.

\textbf{Causal Effect of Concept $c_i$.} Drawing inspiration from existing literature~\cite{feder-etal-2021-causalm}, we formulate the causal effect of concept $c_i$ (for a specific input) as:

\begin{multline}\label{rite}
    \widehat{RITE}_i = < \phi(I^{{CF}_i}_{DC}) - \phi(I_{DC}) > \\
    = < \phi(f((t,v), L_{DC}^{-i})) - \phi(f((t,v), L_{DC})) >
\end{multline}

Here, $\widehat{RITE}_{i}$ represents the \textit{"Representation-based Individual Treatment Effect"}, for $c_i$, with $DC$ as subscript representing the de-confounding objective. The function $f$, modelled as VisualBERT in this paper, takes the input meme as text and visual representation ($(t,v)$) and latent $L_{DC}$, outputting a representation $I_{DC}$. This intermediate representation is then passed through a classifier $\phi$. \textit{Essentially, $\widehat{RITE}_i$ denotes the absolute change in predicted class probability due to the absence of concept $c_i$ and could be used to measure its causal effect on the model.}

\subsection{Concept Annotation}
% As illustrated in Figure \ref{fig:el}, the causal process requires certain concepts along with the meme input which together make the prediction. The set of concepts should contain a minimal number of concepts which should encompass the whole training set in terms of the concepts embedded in the memes. As an example, let us assume the first meme in the training set is about `terrorism', so the concept set would contain the concept `terrorism'. Let's say the second meme is about `holocaust', so the concept set would contain \{`terrorism', `holocaust'\}. If the third meme is about terrorism, the concept set would not change. If the annotator comes across any new concept, he is instructed to append it to the concept set, otherwise, if the concept is already present in the concept set, leave it as it is.

Figure \ref{fig:el} illustrates the causal process, which relies on the integration of concepts alongside meme inputs to facilitate model predictions. The selection of these concepts (which form the `concept set') is pivotal, aiming to i) encapsulate the breadth of themes present within the training dataset while ii) minimizing redundancy. 

\textbf{Scalability.} To make the annotation process efficient and scalable, we use the following approach. Starting empty, the concept set expands as new concepts are introduced. For example, if the first meme includes `terrorism' and `holocaust', these are added to the set. If the second meme includes `terrorism' and `racism', the set becomes {`terrorism', `holocaust', `racism'}. Therefore, we \textit{\textbf{only} append new concepts to the set if they are absent, to minimize redundancy and counter overlap between similar concepts.} 

\textbf{Annotation Process.} We enlisted three annotators, all postgraduate students aged 25-27 with expertise in multimodal machine learning and prior experience curating datasets published in reputable venues, to annotate the concept set. We ensured ethics and took active steps to ensure their well-being, as detailed in Appendix Section \ref{ethics}.  Let us denote the three annotated concept sets as $\{c_1\}$, $\{c_2\}$, and $\{c_3\}$. We then calculate concept representation $r_t(c_1^i)$, where $c_1^i$ denotes $i$th concept from set $\{c_1\}$. Similarly, let us denote meme representation as $r_t(T^j) \odot r_v(V^j)$, where $T^j$, $V^j$ are the text and image of the $j$th meme from the test set, and $\odot$ illustrates element-wise multiplication. Here, $r_t(T^j) \in \mathbb{R}^{1 \times 768}$ and $r_v(V^j) \in \mathbb{R}^{1 \times 768}$ represent CLIP~\cite{radford2021learning} text and vision encodings respectively for the $j$th meme. We then calculate the \textbf{total similarity} of the set $\{c_1\}$ to the memes in the training set as $\sum_{j=0}^N \sum_{i=0}^n r_t(c_1^i)^T \cdot (r_t(T^j) \odot r_v(V^j))$, where $N$ is the number of memes in the training set and $n$ is the number of concepts in the set $\{c_1\}$. Similarly, this total similarity is calculated for sets $\{c_2\}$, and $\{c_3\}$. We observe that the total similarity of set $\{c_2\}$ is the highest which leads us to choose this as the final set. The concepts in $\{c_2\}$ are shown in Table \ref{tab:annot}.

\textbf{Ambiguity Resolution.} Annotators were instructed to maintain precision and leverage existing annotated concepts when annotating new memes.

Despite these guidelines, ambiguities often emerged, especially when new meme concepts were semantically similar to already annotated concepts. In these instances, annotators were encouraged to engage in discussions to reach a consensus if they encountered difficulties in identifying or formulating a concept. Additionally, they were advised to utilize shared resources, such as dictionaries, to identify semantically related terms.

\begin{table*}[t]
\centering
\resizebox{\textwidth}{!}{%
\begin{tabular}{cccccc}
\hline
0. holocaust (535)    & 1. nazism (510)      & 2. genocide (292)  & 3. funny (3000)                & 4. anti-muslim (345)     & 5. terrorism (276) \\
6. violence (360)     & 7. politics (122)    & 8. racism (405)    & 9. international-relation (88) & 10. adult (172)           & 11. gore (178)      \\
12. misogynistic (381) & 13. immigration (430) & 14. extremism (431) & 15. immoral (151)               & 16. white supremacy (205) & 17. indecency (831) \\ \hline
\end{tabular}%
}
\caption{Annotated concepts and the number of memes that have this concept in brackets. Note that the total number of concepts may exceed the number of memes as a meme can have multiple concepts. Annotators could see the train labels while annotating and were instructed to label non-offensive memes as funny.}
\label{tab:annot}
\end{table*}

\section{Methodology}

\begin{figure}[ht]
    % \centering
    \includegraphics[height=5cm,width=\columnwidth]{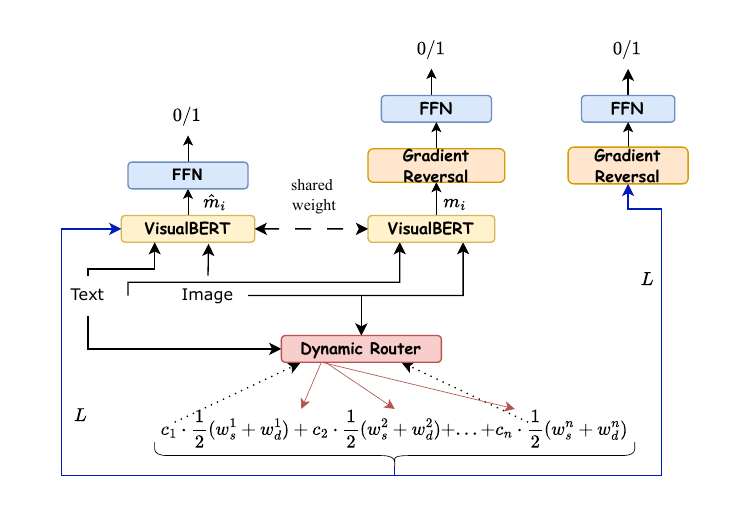}
    \caption{Model architecture comprising of VisualBERT, a dynamic routing layer and a gradient reversal layer. 0/1: non-offensive/offensive.}
    \label{fig:arch}
\end{figure}

% Our method relies on a three-step process, namely i) Modeling with VisualBERT, ii) Adversarial learning and iii) De-confounding.

Our method relies on a three-step process, namely i) Dynamic Routing, ii) Adversarial learning and iii) De-confounding.

% \textbf{Model Inputs.} Considering a meme to be comprised of a text $T$ and an image $I$, we use FasterRCNN to obtain an image feature comprising of the dimension $B \times N \times 768$. Similarly, the text tokens are passed through a BERT embedding layer which outputs text features having dimension $B \times M \times 768$. We concatenate the text features and image features and use this as an input to VisualBERT having dimension $B \times (M+N) \times 768$. Along with that, We use a latent representation $L$ as input having dimension $B \times 1 \times 768$. The latent $L$ is obtained by weighted summation of concept features $c_i$. Each concept features are obtained by passing a textual concept $t_i$ (e.g. ``Holocaust") through a BERT embedding layer. Formally, $L = \sum_{i=0}^{n}{(w_d^i + w_s^i) c_i}$, where $w_d^i$ is a dynamic weight $w_d^i = f(c_i,T)$ and it is a function of $i$th concept $c_i$ and the meme text $T$.
% The functional relationship is learned through dynamic routing as we describe below. Similarly, $w_s^i$ constitutes the static weights which only depend on the concept in question, $w_s^i = f(c_i)$.

\textbf{Model Inputs.} A meme consists of text $T$ and image $I$. We extract image features using FasterRCNN~\cite{ren2016faster}, yielding dimensions $v \in \mathbb{R}^{\mathbb{B} \times \mathbb{N} \times 768}$. Text tokens pass through the model embedding layer to generate text features with dimensions $t \in \mathbb{R}^{\mathbb{B} \times \mathbb{M} \times 768}$. Concatenating these text and image features ($t,v$) results in input dimensions of $\mathbb{R}^{\mathbb{B} \times (\mathbb{M}+\mathbb{N}) \times 768}$ for VisualBERT. Additionally, we introduce a latent representation $L$, with dimensions $\mathbb{R}^{\mathbb{B} \times 1 \times 768}$, obtained by weighted summation of concept features $c_i$. Formally, $L = \sum_{i=0}^{n}{(w_d^i + w_s^i) c_i}$, where $w_d^i$ is a dynamic weight, $w_d^i = \rho(c_i,T)$, dependent on the $i$th concept $c_i$ and the meme text $T$. This functional relationship is learned through dynamic routing. Similarly, $w_s^i$ are static weights, $w_s^i = \tau(c_i)$, dependent only on the concept $c_i$.
% Each concept feature is derived by embedding a textual concept $t_i$ (e.g., "Holocaust") through a BERT layer. 

% \textbf{Dynamic Routing.}

% To model the interaction between $M$ text features $\{u_j\}_{j=0}^M$ output by VisualBERT when given input $T$, and concept feature $c_i$, we learn a weight $W$, which modifies $u_j$ as $u_{ij} = W_{ij} \cdot u_j$. Further, the interaction between $c_i$ and $u_j$ can be modelled by taking a dot product between them. $p_{ij} = u_j^T \cdot c_i$ demonstrates this operation where $p_{ij}$ is a scalar. To normalize the value of $p_{ij}$ between $0$ and $1$, we use the softmax function where, $b_{ij} = \frac{exp(p_{ij})}{\sum_{k=0}^{n}exp(p_{ik})}$, where $n$ is the number of concepts. To measure the effect of all text inputs on concept $i$, we take the weighted mean $s_i = \frac{1}{m}\sum_{j=0}^{m} b_{ij} \cdot u_{ij}$, where $m$ is the number of text tokens input to the model. $s_i$ shows the cumulative effect of all the text inputs on concept $c_i$. We take the intuition that, to model the interaction between $c_i$ and $T$, the length of $s_i$ should approximate their interaction. Specifically, longer vectors should have a unit length whereas shorter vector has a length close to zero. This is mediated by the following squashing function.
% \begin{equation}
%     v_i = squash(s_i) = \frac{|| s_i ||^2}{1+ || s_i ||^2} \cdot \frac{s_i}{||  s_i ||}
% \end{equation}

% The length of $v_i$ then acts as the dynamic weight between $c_i$ and $T_i$, such that, $w_d^i = f(c_i, T_i) = || v_i ||$.

\subsection{Dynamic Routing.} 
\textbf{Need for Dynamic Routing.} The key idea behind dynamic routing is to learn dynamic weights that determine the importance of each concept for a prediction based on both the meme input and the concept itself. These weights are functions of the meme's text, and concept, serving as learnable parameters that control each concept's influence on the prediction. Without dynamic routing, static weights are used, treating all concepts equally, regardless of the specific meme input.

\textbf{How is it achieved?}

To model the interaction between the $M$ text features $\{t_j\}_{j=0}^M$ produced by VisualBERT for a given input $T$ and the concept feature $c_i$, we learn a weight $W$, which modifies $t_j$ as $t_{ij} = W_{ij} \cdot t_j$. Further, the interaction between $c_i$ and $t_j$ can be modelled by taking a dot product between them. $p_{ij} = t_j^T \cdot c_i$ demonstrates this operation, where $p_{ij}$ is a scalar. To normalize $p_{ij}$ between $0$ and $1$, we use the softmax function: $b_{ij} = \frac{exp(p_{ij})}{\sum_{k=0}^{n}exp(p_{ik})}$, where $n$ is the number of concepts. To measure the effect of all text inputs on concept $i$, we calculate the weighted mean: $s_i = \frac{1}{M}\sum_{j=0}^{M} b_{ij} \cdot t_{ij}$, where $M$ is the number of text tokens input to the model. Here, $s_i$ shows the cumulative effect of all the text inputs on concept $c_i$. To model the interaction between $c_i$ and $T$, we want the length of $s_i$ to approximate their interaction. Specifically, longer vectors should have a unit length, whereas shorter vectors should have a length close to zero. This is achieved using the following squashing function~\cite{sabour2017dynamic}:

\begin{equation}
v_i = \text{squash}(s_i) = \frac{|| s_i ||^2}{1+ || s_i ||^2} \cdot \frac{s_i}{|| s_i ||}
\end{equation}

The length of $v_i$ acts as the dynamic weight between $c_i$ and $T_i$, such that $w_d^i = \rho(c_i, T) = || v_i ||$.

% \subsection{Adversarial learning}

% Let us denote $m_i$ a $1 \times 768$ dimensional vector output corresponding to `[CLS]' token when the VisualBERT is presented with the text $T$ and the image $I$ as input. Similarly, let us denote, that $\hat{m_i}$ is the output vector when VisualBERT is presented with the text $T$, the image $I$, and the latent $L$. For classifying an input meme, we use a feed-forward neural network (FFN) on representation $\hat{m_i}$. Normally, input text and images contain enough information to classify an input meme into either offensive or non-offensive classes. That means that the latent representation is not that effective compared to the text and image inputs alone. To make the latent as effective as the text and image, we resort to the use of adversarial learning. Here, the objective is to make both $m_i$ and $L$ invariant to the output class, whereas their combined representation $\hat{m_i}$ should be able to classify an input meme. The purpose is to learn $L$ which should be as much effective as text and image combined. To facilitate this objective, we use a Gradient reversal layer before passing $L$ and $m_i$ to two separate classifiers for the offensiveness detection task. Those two classifiers help to learn class invariant $L$ and $m_i$, whereas the classifier which employs $\hat{m_i}$ learns class-dependent representation.

\subsection{Adversarial Learning}
\textbf{Notation.} Let $m_i$ be a $\mathbb{R}^{1 \times 768}$ dimensional vector output corresponding to the `[CLS]' token when VisualBERT processes the text $T$ and image $I$. Similarly, let $\hat{m_i}$ be the output vector when VisualBERT processes the text $T$, image $I$, and latent $L$ (by concatenating with image representation). For classifying an input meme, we utilize a feed-forward neural network (FFN) on the representation $\hat{m_i}$.

\textbf{Need for Adversarial Learning.} Typically, input text and images contain sufficient information to classify a meme into offensive or non-offensive classes, rendering the latent representation less effective compared to text and image inputs alone.
To enhance the effectiveness of the latent representation to match that of text and image inputs, we employ adversarial learning. The objective is to make both $m_i$ and $L$ invariant to the output class, while their combined representation $\hat{m_i}$ should retain discriminatory information for classifying memes. The aim is for $L$ to be as effective as the combined text and image inputs.

\textbf{How is it achieved?} To achieve this, we utilize a Gradient Reversal Layer~\cite{ganin2015unsupervised} before passing $L$ and $m_i$ to two separate classifiers for the offensiveness detection task. These classifiers help learn class-invariant $L$ and $m_i$, while the classifier utilizing $\hat{m_i}$ learns class-dependent representation.

\subsection{De-confounding}\label{dec}
The premise of de-confounding and its necessity is described in Section \ref{caudiag}. Before de-confounding, we assume, that there exists some $\overline{W}$, which can project each $L^{-i}$ to $c_i$. More formally, $\overline{W} \cdot L^{CF} = C$, where $L^{CF} = [-L^{-1}-, -L^{-2}-, ..., -L^{-n}-] \in \mathbb{R}^{768 \times n}$, and 
$C = [-c_1-, -c_2-, ..., -c_n-] \in \mathbb{R}^{768 \times n}$, where $L^{-i}, c_i \in \mathbb{R}^{1 \times 768}$ and there are $n$ concepts.

% For De-confounding, we want to learn a projection matrix $P$, such that if we project each $c_i$ via it, the resulting counterfactual latents $L^{CF'}$ can never be used to get back $C$. The following theorem demonstrates this.

% \textbf{Theorem.} If $P$ is a nullspace projection matrix of $\overline{W}$, and $C' = P \cdot C \forall i$, then $\overline{W} \cdot L^{CF'} = 0 \forall i$.

% \textbf{Proof.} The latent representation $L'$ after projecting $c_i$ with $P$ is
% \begin{equation}\label{cf}
%     L' = \sum_{i=0}^n{w_i P \cdot c_i} = P \cdot \sum_{i=0}^n{w_ic_i} = P \cdot L
% \end{equation}

% From Equation \ref{cf}, $\overline{W} \cdot L' = \overline{W} \cdot (P \cdot L) = 0$, as $\overline{W} \cdot (P \cdot x) = 0 \forall x$.

% Also as $w_i \neq 0$, and $w_i \overline{W} \cdot (P \cdot c_i) = 0$ and as $\overline{W} \cdot L' = 0$,\\
% $\overline{W} \cdot L^{CF'} = \overline{W} \cdot L' - w_i \overline{W} \cdot (P \cdot c_i) = 0 \forall i$

% Following this theorem and the associated proof, we construct a nullspace projection $P$ of $\overline{W}$. $\overline{W}$ is learned after training a neural network to reconstruct $C$ from $L^{CF} \forall i$. To Deconfound the system, we project the concept representations via $P$, such that, $C \leftarrow P \cdot C$. This idea of finding the nullspace is inspired by Iterative Nullspace Projection (INLP)\cite{ravfogel-etal-2020-null}. After this operation is performed, we denote the latent $L$ with a subscript, i.e. as $L_{DC}$.

For De-confounding, we aim to learn a projection matrix $P$ (a nullspace projection matrix of  $\overline{W}$) such that projecting each $c_i$ through it yields counterfactual latent $L^{CF'}$ incapable of reconstructing $C$. The following theorem illustrates this.

\textit{{\ul Theorem.}} If $P$ is a nullspace projection matrix of $\overline{W}$ and $C' = P \cdot C$ \textit{for all} $i$, then $\overline{W} \cdot L^{CF'} = 0$ \textit{for all} $i$. For \textit{{\ul Proof}}, refer to Appendix Section \S \ref{ap-pr}. After de-confounding, we denote the latent $L$ with a subscript, i.e., $L_{DC}$.

\section{Experimental setups}

The experimental setups and dataset details are elaborated in the Appendix Section \ref{exp-setup} due to space constraints. Here we introduce various metrics and baselines. 

\subsection{Simulating Model Outcome}

\textbf{Definition.} Simulatability, as defined in \citet{hase-etal-2020-leakage}, refers to how well explanations from model $M$ help an observer (e.g. another simpler model, called a simulator model) predict outputs of $M$. Intuitively, better simulatability would reflect faithful explanation as there is a pattern between explanation as input and model prediction as output for the simulator to learn. In our proposed framework, explanation would mean causally sorted concepts in descending order. Sorting is necessary to preserve the causal importance order between the concepts. 

\paragraph{Notation.} Let us assume an ordered set $X^j_{cau} = \{x_i^j\}_{i=1}^n$ contains $n$ concepts sorted by their causal attribution score (measured by $\widehat{RITE}$ score) in a decreasing manner for the $j$th meme in the dataset. Similarly, let us assume an ordered set $X^j_{attr}$ refers to $n$ concepts sorted by their input attribution score by an input attribution method like Integrated Gradient.

\paragraph{Rank Correlation.}
A positive correlation between these sets $X^j_{cau}$ and $X^j_{attr}$ indicates that highly attributed concepts also influence model outcomes (causality), while a negative correlation suggests the opposite. We measure correlation by both Kendal's~\cite{10.1093/biomet/30.1-2.81} and Spearman's rank correlation~\cite{ca468a70-0be4-389a-b0b9-5dd1ff52b33f} criteria.

\paragraph{Accuracy Metrics.} Denoting $\hat{x}_i^j$ as the text representation of $x_i^j$, the rank-adjusted\footnote{rank adjustment is necessary for maintaining the order.} BoW representation $\hat{X}^j$ (note that we drop $cau$ or $attr$ as subscript to show a generic case) of the set $X^j$ would be $ \frac{1}{n} \sum_{i=1}^{n} { \gamma ^ i \hat{x}_i^j}$, where $\gamma=0.9$ is a positive non-zero constant. We train a simulator (a support vector machine-SVM model~\cite{cortes1995support}) on: i) $\hat{X}^j$, ii) concatenation of $\hat{X}^j$ and $\hat{m}_i^j$, denoted by $[\hat{X}^j; \hat{m}_i^j]$, and iii) $\hat{m}_i^j$, to predict the original model prediction $\hat{y}^j$. The $j$th superscript reflects the $j$th meme. Intuitively, the SVM simulates the original model based on the provided information (either one of cases (i), (ii), or (iii)). The performance of the simulator can be seen in Table \ref{tab:my-table1}. \textit{F1 w/ exp} denotes the simulator performance in case (i) when only the ranked concepts $X_{cau}$ or $X_{attr}$ were used as simulator input. Similarly, \textit{F1 w/ inp} denotes case (iii), where the multimodal representation ($\hat{m}_i^j$) is used for model input. Lastly, \textit{F1 w/ both} denotes case (ii). 

\paragraph{Comprehensiveness and Sufficiency.} 
To measure the impact of set $X_{cau}$ or $X_{attr}$ on simulator performance, we employ two metrics: i) Comprehensiveness and ii) Sufficiency~\cite{deyoung-etal-2020-eraser}. Comprehensiveness quantifies the reduction in simulator model confidence when $\hat{X}^j$ replaces $[\hat{X}^j; \hat{m}_i^j]$ as simulator input. Denoting the simulator by $S$, comprehensiveness is $S([\hat{X}^j; \hat{m}_i^j])_k - S(\hat{m}_i^j)_k$ for predicted class $k$. A higher comprehensiveness score indicates more importance of attribution set $X^j$ for the simulator. Sufficiency is defined as $S([\hat{X}^j; \hat{m}_i^j])_k - S(\hat{X}^j)_k$. It requires a higher average comprehensiveness and lower average sufficiency score for $X^j$ to be considered more simulatable.

\subsection{Quantifying Trustworthiness}

To measure the trustworthiness of a model, understanding whether its predictions originate from relevant concepts within the input is crucial, akin to assessing if the model is `right for the right reasons.' To ascertain this, we annotate offensive memes from the test set with relevant concepts from a predefined set of $18$ concepts. Subsequently, we employ averaged Precision@5 (P@5), Recall@5 (R@5), and Mean Average Precision (MAP@5) to assess the relevance of the top five concepts (w.r.t. the annotation) from both the $X_{cau}$ and the attribution set $X_{attr}$ obtained through various input attribution methods. Technical specifics are detailed in the Appendix Section \ref{evmat}. Better scores in all these metrics reflect that the top attributed concepts align with human judgement, thus essentially making the model more trustworthy in return.

\subsection{Baselines} 

% We use various baselines to calculate attribution scores. The baselines we use are standard input attribution based baselines.
% The baselines can be divided into two groups: i) \textit{Path-integral based gradient attribution:} a) Integrated Gradient\cite{sundararajan2017axiomatic}, b) DeepLIFT\cite{shrikumar2019learning}, c) DeepLIFTSHAP\cite{lundberg2017unified}, d) GradientSHAP\cite{ancona2018better}; and ii) \textit{Simple gradient attribution:} a) Saliency\cite{simonyan2014deep} and b) Input X Gradient\cite{shrikumar2017just}. 

% We calculate gradient attribution for all the methods using Captum library (\url{https://captum.ai}). We calculate attributions for three matrices (i.e. groups) separately: i) Concept matrix $C \in \mathbb{R}^{18 \times 768}$, where $18$ concepts are there and each concept is represented by a $\mathbb{R}^{1 \times 768}$ dimensional vector, ii) Textual embedding ($t \in \mathbb{R}^{\mathbb{B} \times \mathbb{M} \times 768}$) and iii) Image embedding ($v \in \mathbb{R}^{\mathbb{B} \times \mathbb{N} \times 768}$).

% Completeness property as satisfied by path-integral based methods have interesting connection with Causality (i.e. how $\widehat{RITE}$ score is calculated), details of which are described in the Appendix Section \ref{cvc}.

We employ several standard input attribution methods to calculate attribution scores, dividing them into two groups based on their underlying mechanisms:

\textbf{Path-integral based Gradient Attribution:}

\begin{itemize}
    \item Integrated Gradients (IG)~\cite{sundararajan2017axiomatic}: IG attributes the importance of features by integrating gradients along the path from a baseline input to the actual input. It ensures that attribution is distributed across all input features in a manner that satisfies the completeness property (\S Appendix \ref{cvc}).
    \item DeepLIFT~\cite{shrikumar2019learning}: This method compares the activation of each neuron to a reference, and assigns importance to the neuron based on the comparison score.
    \item DeepLIFTSHAP~\cite{lundberg2017unified}: A variant of DeepLIFT that aligns with SHAP values, combining both methods to compute attribution based on a cooperative game-theory approach.
    \item GradientSHAP~\cite{ancona2018better}: GradientSHAP samples a point between an input-baseline pair and computes the mean gradients with respect to an output class across all such pairs.
\end{itemize}

\textbf{Simple Gradient Attribution:}

\begin{itemize}
    \item Saliency~\cite{simonyan2014deep}: It Identifies key input features by computing output gradients, highlighting features most influential to the model's prediction.
    \item Input × Gradient~\cite{shrikumar2017just}: This method computes the element-wise product of the input and its gradient to measure each feature's contribution to the prediction.
\end{itemize}

We use the Captum library (\url{https://captum.ai/}) to calculate attributions for each method and apply these attributions to three matrices: i) the concept matrix ($C \in \mathbb{R}^{18 \times 768}$), ii) the textual embedding ($t \in \mathbb{R}^{\mathbb{B} \times \mathbb{M} \times 768}$), and iii) the image embedding ($v \in \mathbb{R}^{\mathbb{B} \times \mathbb{N} \times 768}$).

\textbf{Comparison with Our Methodology.} 
Traditional input attribution methods outlined above rank features (or concepts) by assigning scores based on their importance to the model's output. In contrast, our approach adopts a causal perspective, using the $\widehat{RITE}$ score to measure how much each concept causally influences the prediction. 

While our causal framework differs from standard attribution techniques, we explore an intriguing link between the completeness property of path-integral methods and causality, which we discuss in more detail in Appendix \ref{cvc}.

% While the above baseline methods calculate attributions based on the importance of features for the produced output, our approach takes a causal perspective. In standard input attribution, each feature (or concept) receives a score based on its contribution to the output, and these scores are then ranked to determine feature importance.

% In contrast, we evaluate the causal relevance of each concept using the RITE score, which identifies how much each concept causally impacts the prediction. 

% Although our proposed causal approach and baselines do not match in principle, there is an interesting connection between the completeness property of path-integral methods and its connection to causality, elaborated in Appendix Section \ref{cvc}.

\section{Quantitative Findings}\label{quant}

\begin{table*}[!t]
\tiny
\centering
\resizebox{\textwidth}{!}{%
\begin{tabular}{@{}cccccccc@{}}
\toprule
\multicolumn{1}{l}{}                 & \multicolumn{2}{c}{\textbf{Causal Rank Correlation}}      & \multicolumn{2}{c}{\textbf{Explainability of Simulator}}   & \multicolumn{3}{c}{\textbf{Performance of Simulator}}                  \\
{\ul Methods}                        & \textit{Kendall's Tau} & \textit{Spearman's rho} & \textit{Comprehensiveness} & \textit{Sufficiency} & \textit{F1 w/ both} & \textit{F1 w/ inp} & \textit{F1 w/ exp} \\ \midrule
Int. Grad.                          & 0.017                  & 0.025                   & 0.030                      & 0.089                & 0.66                & 0.62               & 0.53               \\
Saliency     & \textbf{0.859}                  & \textbf{0.916}                   & 0.005                      & 0.116                & 0.63                & 0.62               & 0.42               \\
DeepLift     & -0.001                 & -0.002                  & 0.029                      & 0.021                & 0.67                & 0.62               & 0.65               \\
DeepLiftSHAP & -0.003                 & -0.005                  & \textbf{0.035}                      & \textbf{0.017}                & 0.67                & 0.62               & \textbf{0.68}               \\
GradientSHAP                         & 0.010                  & 0.014                   & 0.008                      & 0.117                & 0.63                & 0.62               & 0.45               \\
Input x Grad & -0.001                 & -0.002                  & 0.034                      & 0.029                & \textbf{0.68}                & 0.62               & 0.67               \\ \hline
\textit{Causal}                      & 1                      & 1                       & 0.002                      & 0.112                & 0.63                & 0.62               & 0.46 

\\ \bottomrule

\end{tabular}%

}
\caption{Table shows an assessment of input attribution-based methods in two directions: i) Their rank correlation with causality, ii) Explanation capability of the extracted keywords as measured by a simulator. Note that here we do not use de-confounding for input-attribution methods as we want to measure the \textit{overall effect} of these concepts on the model outcome (\S list \ref{cvc2})}
\label{tab:my-table1}
\end{table*}

\begin{table*}[!t]
\tiny
\centering
\resizebox{\textwidth}{!}{%
\begin{tabular}{@{}cllllllllllll@{}}
\toprule
\multicolumn{1}{l}{{\ul Input Attr/Causal}} & \multicolumn{3}{c}{\textbf{Full}}                                                                                              & \multicolumn{3}{c}{\textbf{w/o dyn. routing}}                                                                       & \multicolumn{3}{c}{\textbf{w/o adversarial}}                                                                        & \multicolumn{3}{c}{\textbf{w/o deconfounding}}                                                                    \\ \midrule
\multicolumn{1}{l}{}                        & \multicolumn{1}{c}{\textit{R@5}}     & \multicolumn{1}{c}{\textit{P@5}}     & \multicolumn{1}{c}{\textit{MAP@5}}    & \multicolumn{1}{c}{\textit{R@5}} & \multicolumn{1}{c}{\textit{P@5}} & \multicolumn{1}{c}{\textit{MAP@5}} & \multicolumn{1}{c}{\textit{R@5}} & \multicolumn{1}{c}{\textit{P@5}} & \multicolumn{1}{c}{\textit{MAP@5}} & \multicolumn{1}{c}{\textit{R@5}} & \multicolumn{1}{c}{\textit{P@5}} & \multicolumn{1}{c}{\textit{MAP@5}} \\
Int. Grad.                                 & 0.29                                  & 0.20                                  & 0.20                                  & 0.26                              & 0.17                             & 0.19                               & 0.28                             & 0.19                              & 0.17                              &  0.27       & 0.18       &  0.17        \\

Saliency                                    & 0.27           &  0.19           &  0.18           &  0.19       &  0.12       & 0.16        &  0.29       &  0.19       &  0.16        & 0.32       & 0.21       & 0.25       \\

DeepLIFT                                    & 0.30                                  & 0.21                                  & 0.21                                  & 0.27                              & 0.18                             & 0.17                               &  0.26       & 0.18      &  0.16        & 0.26       &  0.18       &  0.18        \\

DeepLIFTSHAP                                & 0.31                                 & 0.21                                 & 0.22                                  & 0.31                              & 0.22                              & 0.22                               & 0.30       & 0.20       &  0.20        &  0.29       & 0.19       &  0.20        \\

GradSHAP                                    & 0.30                                  & 0.20                                 & 0.20                                  &  0.30       & 0.21       & 0.21        & 0.29       &  0.20       &  0.20        &  0.29       & 0.20       &  0.20        \\

Input x Grad                                & 0.30                                  & 0.21                                  & 0.21     & 0.26                              & 0.17                             & 0.17                               &  0.27       & 0.19       &  0.17        &  0.26       &  0.18      & 0.17        \\ \hline

\textit{Causal}                             & \cellcolor[HTML]{FD6864}\textbf{0.26} & \cellcolor[HTML]{FD6864}\textbf{0.18} & \cellcolor[HTML]{FD6864}\textbf{0.17} & 0.21                              & 0.14                              & 0.16                               & 0.20                              & 0.14                              & 0.13                               & 0.30                              & 0.20                              & 0.19                               \\ \bottomrule
\end{tabular}%
}
% \caption{Red border denotes scores obtained by \textit{Causal} framework, i.e. sorting keywords wrt their ICaCE score. The other scores are obtained by other input-attribution frameworks are also shown. The keywords are sorted by their input attribution score. We observe that in our Full method, all methods receive lower score compared to causal score. This means causal keywords are more relevant. Saliency obtains better score, also it has a positive rank correlation to the Causal keywords. This suggests Saliency should be practically used the most. In the ablations (e.g. w/o adversarial), input attribution based methods beat causal method despite negative rank correlation with causal keywords. This shows that relevant keywords (as predicted by input attribution) need not always be the cause of prediction, showing the model is mostly right due to the wrong reason when ablating components (shown in red). All in all, this suggests our full framework retrieves causal, relevant and understandable keywords. If we want trade-off, we should go for Saliency based keywords (shown in blue). Saliency based keywords are relevant (as shown here), causal (+ve rank correlation with causal keywords), and explainable (higher LAS than causal keywords), showing that its best of the all world.}
\caption{Red border denotes scores obtained by \textit{Causal} framework, i.e. sorting keywords with their $\widehat{RITE}$ score. The scores from alternative input-attribution methods are also displayed.}
\label{tab:my-table2}
\end{table*}

\paragraph{1. Classifier performance.}Model performance on the test set is $70.36\%$ as measured by the F1 score. This reflects the VisualBERT in our framework works well and even exceeds some of the benchmark models evaluated as a part of the FB Hateful Meme Dataset \cite{kiela2021hateful}. 

\paragraph{2. Simple attribution methods align better with Causality.}
Table \ref{tab:my-table1} presents Kendall's tau and Spearman's rho (averaged across all memes in the test set) for correlation comparison. Notably, simpler methods like Saliency show a stronger correlation with the causal set than complex methods like DeepLIFT, suggesting their potential to capture causal relationships. This suggests that simpler methods may offer a clearer and more direct understanding of causality within the model's decision-making process.

\paragraph{3. Complex attribution methods are more simulatable.}
Consider the \textit{F1 w/ exp} scores for DeepLIFT, DeepLIFTSHAP, and Input x Grad as input attribution methods (\S Table \ref{tab:my-table1}). The attribution set $X_{attr}$ obtained by these methods achieves $\sim 66\%$ F1 score of the simulator, indicating a high correlation between $X_{attr}$ and the original model prediction $\hat{y}$ (termed as highly simulatable). Also both comprehensiveness and sufficiency scores are higher for these models.

\paragraph{4. Correlation does not imply causation.}
Note the negative correlation of $X_{attr}$ with the $X_{cau}$ set for complex attribution methods (\S Table \ref{tab:my-table1}), highlighting the fact that \textit{Correlation (simulation) does not always imply Causation}. Additionally, observe the lower \textit{F1 w/ exp} score obtained by the Saliency attribution set, despite its higher rank correlation to the \textit{Causal} set. This suggests that keywords causally related to model outcome may not always be easily simulatable. This outcome is in line with the study of \citet{bastings-etal-2022-will}, where authors showed that simple attribution methods are more faithful compared to complex methods. \textit{Note that although the findings match, their approach is non-causal and does not overlap with ours at all.}
% The faithfulness of an attribution method depends on whether the ranked concepts reflect causality rather than simulatability. Therefore, Saliency might be important in faithfully interpreting a model, despite its lower simulatability performance.

\begin{figure}
    \centering
    \includegraphics[height=4cm,width=\columnwidth]{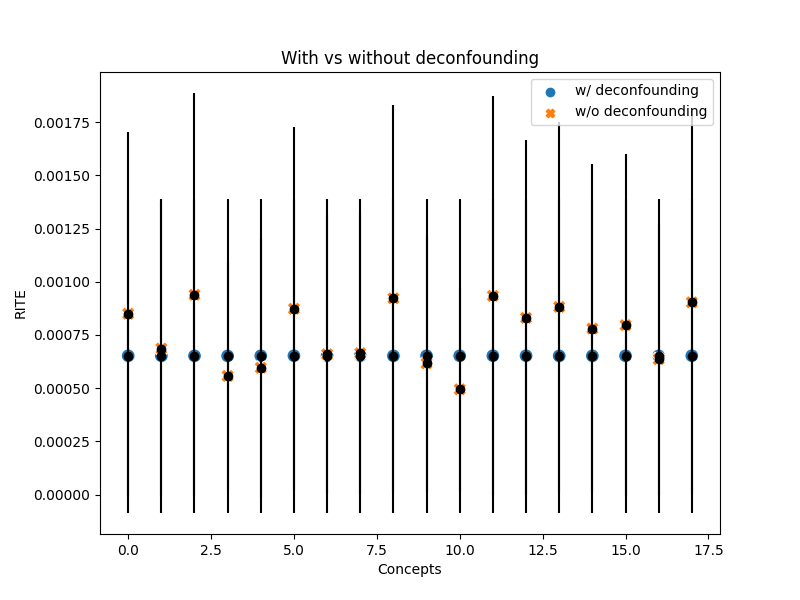}
    \caption{Comparison of mean $\widehat{RITE}$ score between w/ de-confounding and w/o de-confounding strategies}
    \label{fig:comp}
\end{figure}

\begin{table*}[!t]
\centering
\tiny
\resizebox{\textwidth}{!}{%
\begin{tabular}{@{}ccccc@{}}
\toprule
{\ul Meme Idx} & \textbf{Top-5 Causal}                                                                      & \textbf{Actual} & \textbf{Predicted} & \textbf{Possible Reason} \\ \midrule
32                & `anti muslim', `terrorism', `genocide', `violence', `adult'                     & Non Offensive   & Offensive          & Dataset bias             \\

93 & `violence' `racism', `genocide', `anti muslim', `holocaust' & Non Offensive   & Offensive          & Dataset bias             \\

92                & `violence', `genocide', `terrorism', `extremism', `nazism'                                 & Non Offensive   & Offensive          & Inadequate context       \\
65                & `indecency', `holocaust', `funny', `immigration', `racism' & Offensive       & Non Offensive      & Background knowledge     \\
73                & `holocaust', `violence', `gore', `nazism', `funny'                                         & Offensive       & Non Offensive      & Modality conflict        \\ 

76 & `misogynistic`, `gore',  `holocaust', `extremism', `nazism' & Offensive & Non Offensive & Bias and modality conflict \\

\bottomrule
\end{tabular}%
}
\caption{Meme index with corresponding causal keywords and their possible reason behind error cases.}
\label{tab:error-analysis}
\end{table*}

\paragraph{5. Dynamic routing is paramount.}
Within our framework (\S Table \ref{tab:my-table2}), without any ablation of modelling components (shown by Full), the $X_{cau}$ set achieves $0.26$, $0.18$, and $0.17$ scores for R@5, P@5, and MAP@5, respectively. Disabling dynamic routing (w/o dyn. routing) results in static weighting of concepts irrespective of meme input, leading to reduced scores for the $X_{cau}$ set across all metrics compared to the Full framework, indicating diminished trustworthiness.

\paragraph{6. De-confounding is necessary for establishing correlation between causal concepts and model output.}
Without de-confounding, input attribution-based methods like DeepLIFTSHAP and GradientSHAP do not perform as well as they do with other configurations (e.g. w/o dyn. routing). Refer to Table \ref{tab:my-table2}. This suggests that without de-confounding, the model struggles to find a proper correlation between causal concepts and model output. Empirically, in Figure \ref{fig:comp}, we illustrate the mean $\widehat{RITE}$ scores of the concepts with and without de-confounding. Without de-confounding, due to widely varied mean $\widehat{RITE}$ scores for several concepts, certain concepts dominate the top positions in the causal set, regardless of the input, indicating potential bias in the model's causal attribution and making it difficult for input attribution methods to establish a correlation between model input and output, resulting in lower scores across metrics.

\paragraph{7. Necessity of de-confounding for causality.}
The observation that the `causal' set performs better without de-confounding may raise questions about its necessity. However, as illustrated in Section \ref{dgp}, de-confounding remains a principally valid choice. Without de-confounding, as discussed in the previous paragraph, certain concepts often dominate the top positions in the causal set, regardless of the input. Conversely, with de-confounding, all concepts have similar $\widehat{RITE}$ values and their standard deviation, ensuring equal representation across test set examples (\S Figure \ref{fig:comp}). Therefore, despite achieving lower scores, the framework with de-confounding is more trustworthy due to its consistent performance, lack of bias towards certain concepts in the causal attribution set, and equal representation of concepts across inputs.

\begin{figure*}[t]
    \centering
    \includegraphics[width=\textwidth,height=3.4cm]{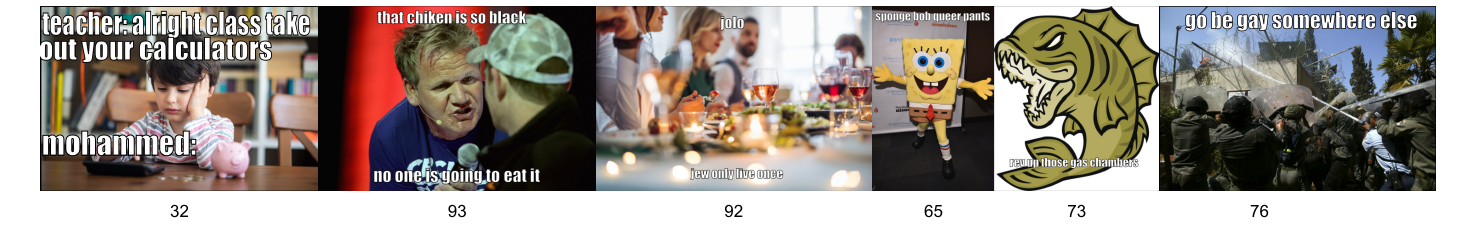}
    \caption{Memes of Table \ref{tab:error-analysis}.}
    \label{fig:error_analysis}
\end{figure*}
\section{Analysing the model through Causal Lens}\label{quals}
\begin{figure}[!ht]
    \centering
    \includegraphics[width=\columnwidth]{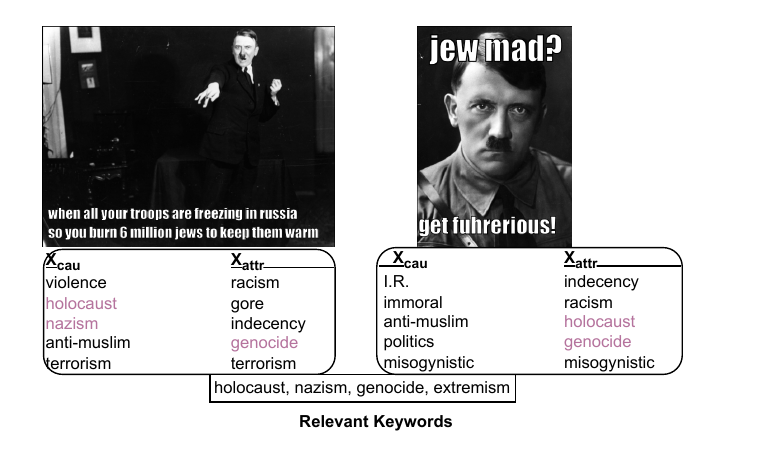}
    \caption{Evaluating trustworthiness of the model}
    \label{fig:trust}
\end{figure}
\textit{Is the model always right due to the right reason?} No, it is not always the case. We gain insight into this by examining the model causally. In Figure \ref{fig:trust}, we present two examples of memes from the test set along with the model's predictions, both correctly classified as offensive.

In the first example, the meme's offensiveness is linked to the Holocaust and antisemitic ideas, accurately reflected in the Top-5 causal concepts. Notably, concepts like `Violence', `Holocaust', and `Nazism' from the gold standard set directly relate to the meme's context. With two out of the Top-5 causal keywords aligning with the meme's context, we conclude that the model's prediction was correct for the right reason.

In the second example, despite the meme being antisemitic, the identified causal concepts are irrelevant to antisemitism, although highly offensive. However, certain keywords (such as `Holocaust' and `Genocide') identified by GradientSHAP are attributed. Three of these (`Racism', `Holocaust', and `Genocide') directly relate to the meme's context. Initially, it may seem the model was right for the right reason. However, the absence of overlap between the set of causal concepts and the gold standard concept set indicates that the model classified the meme correctly but with erroneous causal attribution. This discrepancy suggests that although accurate input attributions exist, the model may base its decision on different causal concepts. Such analyses can help a user to trust the model predictions. 

\section{Error Analysis}

% add more

In this section, we analyze the model's performance using causal concepts. For example, in the first row of Table \ref{tab:error-analysis} (meme index $32$), the model identifies `anti-muslim' among the top 5 causal keywords, even though they are irrelevant. The presence of `Mohammed' alone leads the model to classify the meme as offensive, indicating a dataset-wide bias towards words like `Mohammed'. Similarly for the second meme, the mere presence of the concept of `blackness' was sufficient for classifying the meme as offensive despite having nothing related to racism in it. This also shows dataset bias where lots of racists ($\sim 47.1\%$) and religious memes ($\sim 39.3\%$) are present. 

Similarly, in the third meme, the word `jew' prompts the model to associate the meme with concepts like `violence', and `genocide' possibly due to insufficient visual context and association of the word jew in antisemitic offensive memes.\footnote{\textcolor{red}{The biases identified in the model's behavior do not reflect the views or biases of the authors. The analysis aims to highlight the model's limitations and the importance of addressing bias in machine learning systems.}}

In the fourth meme, although offensive, the model finds it humorous due to a lack of background knowledge, especially regarding the wordplay on SpongeBob, and the smiling SpongeBob face may further contribute to misclassification.

The fifth meme's misclassification can be attributed to a humorous background image unrelated to the meme text, creating a \textit{modality conflict}. Note that the fourth and fifth memes which are classified as non-offensive have the `funny' keyword as a part of the Top 5 causal keywords. 

The sixth meme shows a sexually offensive remark but due to conflicting visual modality (showing violence), this meme got misclassified as non offensive. The keywords generated can also be attributed to dataset bias as shown in the first two memes.

By examining causal keywords alongside model inputs, this type of error analysis offers insights into why the model made mistakes.

\section{Conclusion}

In this paper, we introduce a multimodal causal framework aimed at transparently analyzing VisualBERT predictions. Guided by an SCM, the framework compels VisualBERT to base its decisions on interpretable human-understandable concepts. Evaluation on the Facebook hateful meme dataset reveals key insights: i) Input attribution methods may lack causal underpinning, ii) Modelling choices significantly influence relevant causal attributions, enhancing model trustworthiness. The qualitative analysis delves into whether the model is `right for the right' reasons and uncovers causal factors behind misclassifications.

The simplicity and versatility of our framework (i.e. the underlying Structural Causal Model and its translation to modelling choices) allow its application across various tasks and multimodal models. Although we show the importance of our architecture on meme offensive detection tasks as a testbed, its application may be important in medicine where the need for trustworthy systems is paramount. 
% Its potential extends to domains like medicine and warfare, where safety is paramount. 
% although we showed the importance of our architecture on meme offensive detection task as a testbed, its application may be important in medicine or warfare where trustworthy systems are paramount. Future research may explore its generalizability to diverse fields, paving the way for its broader adoption.

\section*{Limitations}

While our approach demonstrates promising results, there are some limitations to consider. Firstly, the reliance on a specific dataset, such as the Facebook Hateful Meme dataset, and a specific model, like VisualBERT, may limit the generalizability of our findings to other datasets and models.

Secondly, the concept annotation process introduces challenges as it relies on human annotators to define and refine the concept set. This process may introduce subjectivity and biases. To address this challenge, employing more robust annotation guidelines, inter-annotator agreement assessments, and sensitivity analyses can enhance the reliability of the concept annotation process.

At the outset, the concept of the paper acts as a seed or proof of concept, further generalizability of which is to be explored through a chain of related future studies. Specifically, exploring potential applications of the framework in other domains beyond meme classification would be valuable. The framework could be applied in areas such as content moderation, sentiment analysis,  and trend analysis in social media, news media, marketing, and public opinion research and medicine.

\section*{Ethical Declaration}

We acknowledge the potential for misuse of our annotated concepts, which could be employed to filter memes based on racial prejudices. To mitigate this risk, human moderation and intervention are crucial. The purpose of annotating concepts is to facilitate research into the analysis and undertstanding of offensive memes on the internet. When used appropriately, we believe it serves as a valuable resource.

Further the involvement of annotators to annotate potentially triggering meme may seem problematic. On the other hand, we completely ensure annotators' well-being by making voluntary free session with institutional counsellors available at any time. Also, participation in this process was purely at their own wish and they have been warned on exposing themselves to various offensive and trigerring contents which were marked as a disclaimer. We followed four broad ethical principles during the annotation process: i) Annotators were fully briefed on the nature of the task and provided informed consent to participate. ii) Annotators had access to psychological support via our institutional counseling system. iii) Annotators were compensated fairly in line with institute regulations. iv) The privacy and confidentiality of student participants were strictly protected throughout the study.

By adhering to these protocols, we ensured that the ethical concerns associated with using students to label offensive memes were adequately addressed, prioritizing their well-being and ethical treatment. Our study underwent evaluation and approval by our Institutional Review Board (IRB) before proceeding for either annotating offensive memes in the first place or using students to annotate these memes.

\section*{Acknowledgements}
Dibyanayan Bandyopadhyay acknowledges Prime Minister's Research Fellowship (PMRF). The authors further acknowledge anonymous reviewers for their useful feedback.

\bibliography{custom}

\appendix

\section{Completeness vs. Causality}\label{cvc}

Completeness is a property inherent to path-integral-based methods. Consider a concept defined by $c_i$, where $C = \{c_i\}_{i=1}^{n}$ and $n=18$ represents the number of concepts. The completeness property asserts that the attribution for $c_i$, denoted as $Attr_i$, when calculated separately from other concepts, is equal to $F(c_i, (t,v)) - F(c_i', (t,v))$. Here, $F$ is the VisualBERT combined with the classifier, $(t,v)$ represents other unchanged model inputs (meme content), and $c_i'$ denotes the baseline, often chosen as a zero vector.

\begin{equation}
Attr_i = F(c_i, o) - F(c_i', o)
\end{equation}

In this context, $F(x) = (\phi \circ f)(x)$. Choosing $c_i$ as the zero vector makes the attribution exactly equal to the $\widehat{RITE}_i$ score for concept $c_i$ (see Equation \ref{rite}).

Using approximations (e.g., Gauss-Legendre approximation) in calculating the path integral results in $Attr_i \approx \widehat{RITE}_i$, implying that $X_{cau} = X_{attr}$. This has an interesting implication:

\begin{enumerate}
\item $X_{cau} = X_{attr}$ suggests that path-integral-based attribution methods exhibit causal behavior, which is a highly desirable property.
\end{enumerate}

However, achieving this property requires certain assumptions and modeling choices, outlined below:

\begin{enumerate}\label{cvc2}
\item The background SCM must be chosen as a modeling choice, resulting in the same outcome as using a concept as a zero vector and setting its corresponding static and dynamic weights to zero.
\item We must assume that individual concept attributions are calculated separately, independent of other concepts, with separate zero-baselines for each concept. This is not the usual practice, as attributions are typically calculated with text, vision, and concepts considered as three singular inputs to the model with random baselines. Consequently, all concept embeddings ($C = \{c_i\}_{i=1}^{n}$) are treated as a single input for attribution calculation.
\end{enumerate}

\section{LLMs/VLMs Self-consistency Checks}\label{scc}

\textbf{Self-consistency checks.} Self-consistency checks~\cite{madsen2024selfexplanations} are methods used to verify the faithfulness of a model's explanation by ensuring that the model's behavior aligns with expected outcomes. In the context of counterfactual explanations, these checks involve modifying the input so that the model predicts the opposite label. By re-evaluating the prediction with this modified input, we can confirm whether the model produces the expected opposite outcome. If it does, the counterfactual explanation is considered faithful. This process is crucial when using an instruction-tuned language model in a conversational setting, where re-evaluation should occur in a new chat session to avoid bias from prior prompts.

% \textbf{LLM Self-consistency checks.} In the context of LLMs, we define two self-consistency checks in Figure \ref{fig:scc-lm1} and Figure \ref{fig:scc-lm2}. In first session of LLM, denoted as Session 1, we query the LLM with the following template: ``I have a meme which can be described as {caption}. The meme text reads: {meme text}. Is this meme offensive? Only answer Yes/No.". The LLM answers either Yes/No to this question. We then ask why is that meme offensive or non-offensive. Subsequently the LLM is prompted to make a counterfactual of the meme text such that it would answer opposite to what it already predicted regarding whether the meme is offensive or non-offensive. The generated counterfactual text is then embedded into the template and sent to the LLM as counterfactual query. It should answer opposite to what it already predicted as the label of the meme, because it is a purported counterfactual as per the model. In Figure \ref{fig:scc-lm1} and Figure \ref{fig:scc-lm2} we denote two failure cases in such counterfactual self-consistency checks for LLMs. This denotes that LLM explanations are not faithful and should not be trusted to reason about model's inner working although they look very plausible.

\begin{figure*}[!t]
    \centering
    \includegraphics[height=14cm,width=\textwidth]{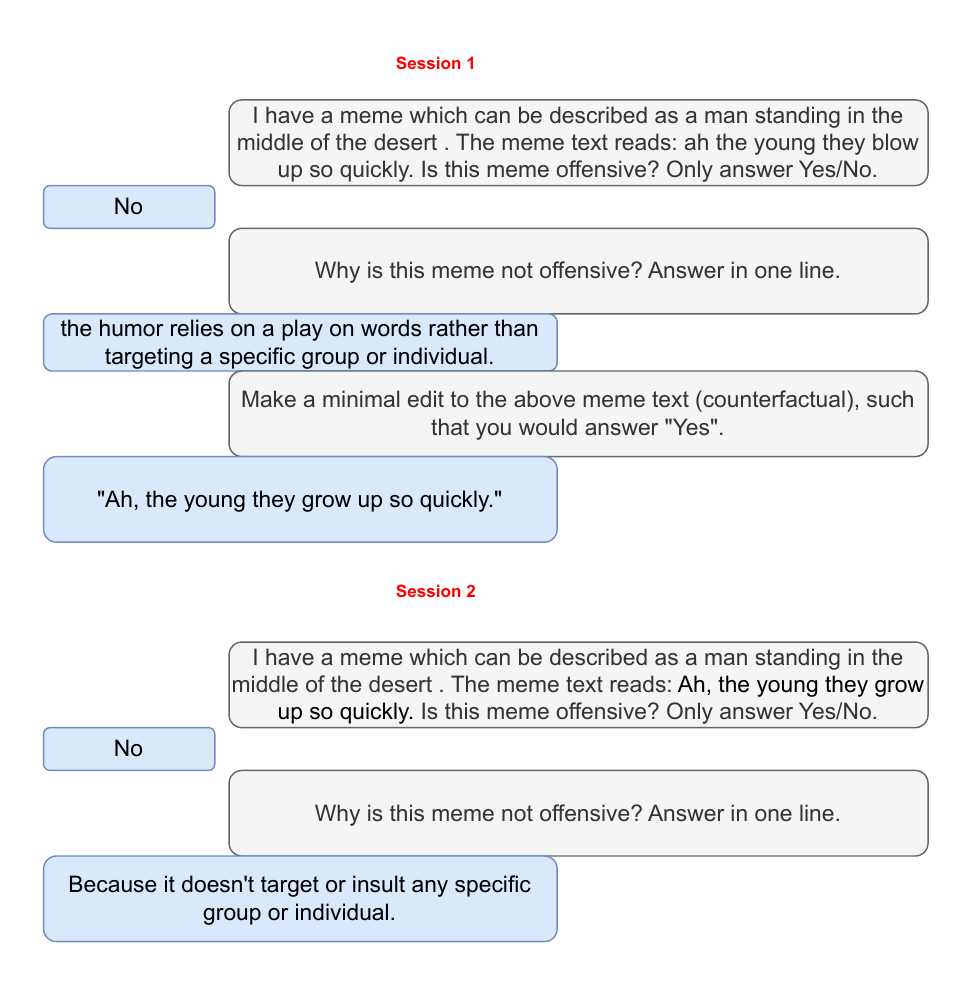}
    \caption{An instance where GPT-3.5 offers an counterfactual self-explanation and employs self-consistency verification to assess its faithfulness.}
    \label{fig:scc-lm1}
\end{figure*}

\begin{figure*}[!t]
    \centering
    \includegraphics[height=14cm,width=\textwidth]{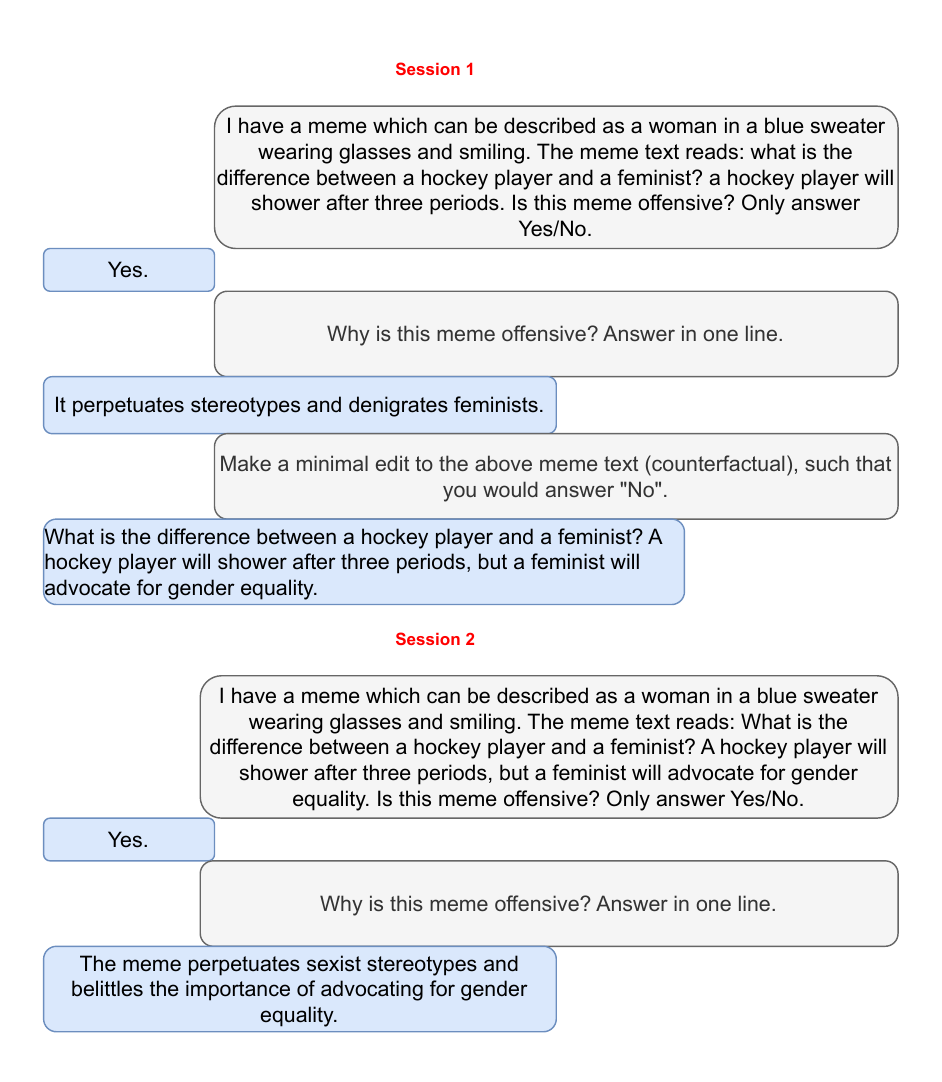}
    \caption{Another instance where GPT-3.5 offers an counterfactual self-explanation and employs self-consistency verification to assess its faithfulness.}
    \label{fig:scc-lm2}
\end{figure*}

\textbf{LLM self-consistency checks.} We define two self-consistency checks for LLMs as illustrated in Figure \ref{fig:scc-lm1} and Figure \ref{fig:scc-lm2}. In the first session, referred to as Session 1, we query the LLM with the following template: "I have a meme which can be described as {caption}. The meme text reads: {meme text}. Is this meme offensive? Only answer Yes/No." The LLM responds with either Yes or No. We then ask the LLM to explain why the meme is offensive or non-offensive. Next, we prompt the LLM to create a counterfactual version of the meme text that would yield the opposite prediction about whether the meme is offensive or non-offensive. This counterfactual text is then embedded into the template and sent back to the LLM as a counterfactual query. The LLM should respond with the opposite label from its initial prediction, as it is a counterfactual according to the model. Figures \ref{fig:scc-lm1} and \ref{fig:scc-lm2} show two failure cases in these counterfactual self-consistency checks for LLMs, indicating that LLM explanations are not faithful and should not be trusted to reveal the model's inner workings, despite their plausible appearance.

% \textbf{VLM self-consistency checks.} Similar to LLMs, Visual Language models (VLMs) can also be used to check for self-consistency. In Figure \ref{fig:cscc-vlm}, we show two examples for VLM self-consistency failure cases using counterfactual explanation. We directly prompt the VLM with the meme to label it as offensive/non-offensive. Then we query the same VLM to generate a counterfactual meme text which is then in-painted again in the meme to generate a counterfactual sample. That counterfactual sample is used as a query in the next session. Being a counterfactual sample to the original one, the original label should flip. But it is not so for both of the cases shown as a part of Figure \ref{fig:cscc-vlm}.

\begin{figure*}[!t]
    \centering
    \includegraphics[height=15cm,width=\textwidth]{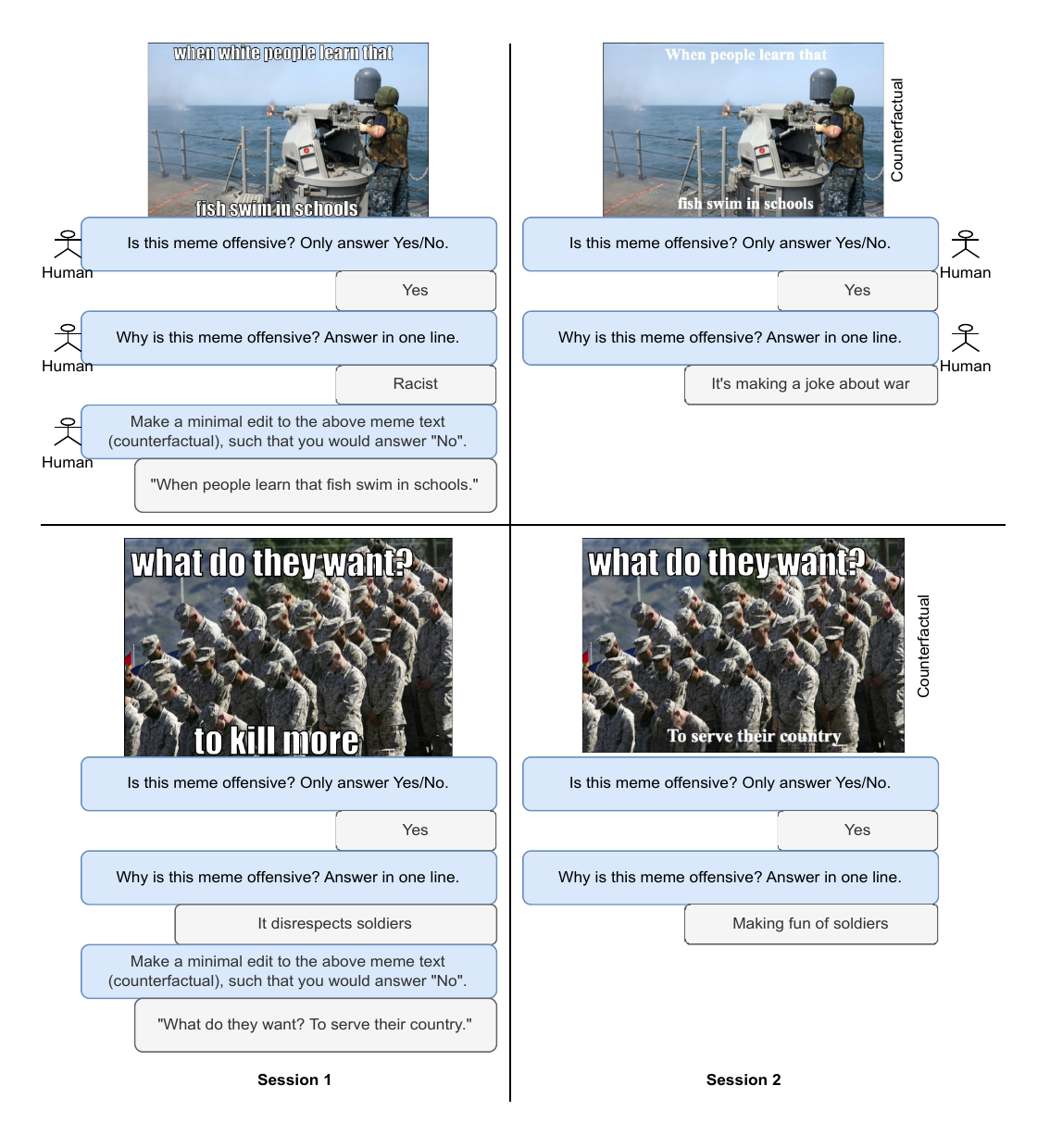}
    \caption{Counterfactual self-consistency checks for VLM, LLaVA-1.6-34B (\url{https://github.com/haotian-liu/LLaVA}).}
    \label{fig:cscc-vlm}
\end{figure*}

\textbf{VLM self-consistency checks.} Similar to language models, Visual Language Models (VLMs) can be assessed for self-consistency. Figure \ref{fig:cscc-vlm}
presents two instances of self-consistency failures in VLMs using counterfactual explanations. Initially, the VLM is prompted to label a meme as offensive or non-offensive. Subsequently, the same VLM is asked to generate a counterfactual version of the meme text, which is then in-painted back into the meme to create a counterfactual sample. This counterfactual sample is used as a query in the following session. As a counterfactual, the original label should be reversed. However, as demonstrated in Figure \ref{fig:cscc-vlm}, this label reversal does not occur in either case, indicating a failure in self-consistency checks for the VLM.

\section{Proof} \label{ap-pr}
The latent representation $L'$ after projecting $c_i$ with $P$ is given by
\begin{equation}\label{cf}
L' = \sum_{i=0}^n{w_i P \cdot c_i} = P \cdot \sum_{i=0}^n{w_ic_i} = P \cdot L
\end{equation}

From Equation \ref{cf}, $\overline{W} \cdot L' = \overline{W} \cdot (P \cdot L) = 0$, as $\overline{W} \cdot (P \cdot x) = 0$ for all $x$.

Furthermore, $w_i \neq 0$, and $w_i \overline{W} \cdot (P \cdot c_i) = 0$. Therefore, $\overline{W} \cdot L^{CF'} = \overline{W} \cdot L' - w_i \overline{W} \cdot (P \cdot c_i) = 0$ for all $i$.

Following this theorem and its proof, we construct a nullspace projection $P$ of $\overline{W}$. $\overline{W}$ is learned by training a neural network to reconstruct $C$ from $L^{CF}$ for all $i$. To de-confound the system, we project the concept representations via $P$, updating $C$ as $C \leftarrow P \cdot C$. This approach is inspired by Iterative Nullspace Projection (INLP)\cite{ravfogel-etal-2020-null}. After this operation, we denote the latent $L$ with a subscript, i.e., $L_{DC}$.

\section{FAQ}

\textbf{1. Question:} \textit{What is the need of the framework?}

Ans: The proposed framework incorporates a VisualBERT and forces it to base its prediction on causal concepts. This makes the VisualBERT transparent because now we can base its decision on the external causal concepts by looking at their $\widehat{RITE}$ scores. As proposed in the Introduction, input memes contain a lot of implicit contexts, which cannot be attributed by Input attribution methods as they are missing from the meme input itself.

\textbf{2. Question: }\textit{What is the need for Adversarial Learning?}

Adversarial Learning ensures the model utilizes the concept alongside the meme inputs. Normally, the meme inputs contain enough information to classify the input meme and simply adding concept representation on top of it does not make the multimodal information any richer. Adversarial learning ensures we learn a non-trivial interaction between multimodal representation and concept representation such that both play an equal role in the final classification task.

\textbf{3. Question: }\textit{With De-confounding, the possibility of top-5 causal concepts being relevant decreases than without it. Then what is its use?}

De-confounding is principally valid because from the counterfactual representation of concept $c_i$, we should not be able to recover $c_i$ (i.e. its representation). Otherwise, it is not even a counterfactual. Our main goal is to estimate an unbiased value of $\widehat{RITE}$ score, which is obtained after de-confounding. So, although w/o de-confounding scores are higher, they may not reflect true causality as without de-confounding, true counterfactual representations are not generated. Also, an empirical argument is validated in the Section \ref{quant} and Figure \ref{fig:comp}.

\textbf{4. Question: }\textit{Why do you call your framework causal when the only thing `causal' is how you select the concepts based on their $\widehat{RITE}$ score?}

We assumed that a meme is offensive due to the meme content itself and some contribution from a fixed collection of concepts. That is illustrated as a structural causal model (SCM) in Figure \ref{caudiag}. Note that every modelling choice is based on the SCM itself. Essentially the framework encompasses an SCM which implements itself as an addition to VisualBERT in the form of i) Adversarial Learning, ii) Dynamic Routing and iii) De-confounding. This forces the SCM to base its decision on meme content as well as on interpretable concepts. Essentially the concepts are called "causal" because they are being modelled as causal in the SCM and the framework implements the SCM on top of VisualBERT.

\textbf{5. Question: }\textit{What is the generality of the proposed framework?}

The framework is overall generic. The framework which implements the SCM is general for every task where implicit domain knowledge is necessary. One such task is meme offensiveness detection. Although the scope of this framework is general, we use it on top of meme offensive detection tasks because of its i) generality of using nontrivial implicit context which is unavailable in input space, and ii) The task has social good applications, and therefore is important to solve. We leave it in the hands of future researchers to use this framework as a proof of concept to describe its effect in critical domains, such as medicine, where the trustworthiness of machine learning (ML) systems is of utmost importance.

So in summary, the proposed framework is generic whose applicability is shown in one interesting task. The applicability of it to other domains remains as an avenue for further research.

\textbf{6. Question: }\textit{In the proposed SCM (Equation \ref{rite}), when a concept is intervened or masked, its information in the text and image remains, potentially affecting the reliability of the causal effects. How do you address this concern?}

We address this issue in two ways: a) \textbf{Gradient Reversal (GR) Layer:} It ensures that the text and image alone, without the explicit presence of the concept, cannot classify the meme as offensive or non-offensive. It forces the model to rely on the explicit presence of the concept for classification.

b) \textbf{De-confounding Strategy:} It nullifies the effects of any concept on other interrelated concepts and provides a clearer analysis of each concept's individual impact on predictions.

These mechanisms ensure the reliability of the causal effects derived from equation (1).

\textit{\textbf{7. Question:} The performance of $X_{cau}$ in "explainability of simulator" and "performance of simulator" is not good compared to $X_{attr}$. Is that problematic?}

It's important to note that evaluating the causal framework based on
simulator scores can be misleading. We are using the causal framework to judge how causally important the input attribution based methods are, not the other way around. Better simulator performance of $X_{attr}$ indicates that it has better recognizable patterns, suggesting it can explain the model's decision well.
Lower scores only on simulatability metrics indicate that $X_{cau}$ does not exhibit patterns like  $X_{attr}$. On the flip side, despite high explainability, input attribution methods do not necessarily capture the underlying causality as indicated by lower rank correlation among various highly simulatable input attribution methods (e.g. DeepLIFT, DeepLIFTSHAP). As noted in this paper, ``Correlation does not imply causation."

\textit{\textbf{8. Question:} Why most of the examples in the paper are related to racism? Is it because offense related to racism are easy to see?}

This is not the case. Most of the examples are racist or religious because they incorporate most of the memes in the dataset, comprising of $47.1\%$ and $39.3\%$ of memes respectively.
\section{Evaluation Metrics}\label{evmat}
\textbf{P@5:} This is defined as among the top 5 causal keywords, how many are also relevant. This can be mathematically formulated as $\frac{n(A \cap B)}{n(B)}$, where $A$ is the set of relevant gold standard keywords, whereas $B$ is the set of top5 causal keywords, and $n(A)$ shows the number of elements in $A$. $n(B)=5$, as we are considering the Top 5 keywords. Consider the first example in Figure \ref{fig:trust}. $A = \{\text{holocaust}, \text{nazism}, \text{genocide}, \text{extremism}\}$ and $B = \{\text{violence}, \text{holocaust}, \text{nazism}, \text{anti-}\\ \text{muslim}, \text{terrorism}\}$. $A \cap B = \{\text{holocaust}, \text{nazism}\}$, which entails $n(A \cap B) = 2$ and $n(B)=5$. So the P@5 score is $2/5 = 0.4$.
We report the averaged P@5 scores across all test set examples in Tables.

\textbf{R@5:} This is defined as among the gold standard relevant keywords, how many are chosen from the  top5 causal set. Following the previous notation, this can be defined as $\frac{n(A \cap B)}{n(A)}$. For the previous example, $n(A \cap B) = 2$ and $n(A)=4$, which entails R@5 is $2/4 = 0.5$. 

\textbf{MAP@5:} This metric is defined as a mean of P@k where k ranges from $1$ to $5$. P@k values are calculated following the previous paragraph. Mathematically, MAP@5 = (P@1+P@2+P@3+P@4+P@5)/5.

\section{Experimental Setup}\label{exp-setup}

Our proposed model was implemented using PyTorch, a Python-based deep-learning library. All experiments (excluding the baselines) were performed on a single Nvidia P100 16GB GPU on CuDA driver version 12.4. The baselines were run on a different CuDA version compared to the baselines ($12.1$ vs $12.4$). We employed the Adam optimizer~\cite{kingma2017adam} with a learning rate set to $5e-5$ for optimization. The experiments were conducted using the Facebook Hateful Meme dataset~\cite{kiela2021hateful}. The whole set of $8500$ memes is broken down into $7840$, $660$  examples as the train-test split using a random seed $42$. The total number of parameters of the model is $350$M and to train a single epoch, $30$ mins of continuous GPU execution is needed with a fixed batch size of $32$. 

To mitigate the impact of non-deterministic GPU operations, we used a fixed random seed of $42$ across all the experiments.

Note that many attribution methods, such as saliency, rely on gradient backpropagation (particularly the chain rule) to compute attribution scores. Small precision differences caused by changes in GPU hardware can lead to significant variations in these scores.

\section{Addressing Ethical Concerns Related to Annotation of Offensive Memes}\label{ethics}
The involvement of annotators in labeling potentially triggering memes may seem problematic. However, we have implemented several measures to ensure the well-being of the annotators:

\textit{Voluntary Participation and Informed Consent:} Participation in the annotation process was entirely voluntary. Annotators were fully informed about the nature of the task and the potential exposure to offensive and triggering content through a detailed disclaimer. They provided their informed consent before participating.

\textit{Psychological Support:} We ensured that annotators had access to free, voluntary sessions with institutional counselors at any time. This support system was established to address any psychological distress that might arise from the task.

\textit{Fair Compensation:} Annotators were fairly compensated for their work, receiving rates according to institutional policies. This compensation was provided in addition to the fellowships that students receive, in accordance with institute regulations.

\textit{Confidentiality:} We maintained strict confidentiality and privacy for all student participants throughout the study.

\textbf{Ethical Review and Approval}: Our study, including the use of students to annotate offensive memes, underwent evaluation and received approval from our Institutional Review Board (IRB). The IRB reviewed our study design, the nature of the content, and the involvement of student participants to ensure compliance with ethical standards.

\textbf{Use of Publicly Available Dataset}: The memes included in our study were selected from the well-known, publicly available FB Hateful Meme Dataset. This dataset represents a wide range of themes found on the internet, including some offensive and extreme content. Our goal was to analyze how different types of content are perceived and categorized by labeling them with keywords, necessitating the inclusion of a broad spectrum of examples.

\end{document}